\documentclass[a4paper,conference]{IEEEtran}
\ifCLASSINFOpdf
 \usepackage[pdftex]{graphicx}
\else
\fi
\usepackage{amsmath}

\begin{document}
\title{Deep Localization of Mixed Image Tampering Techniques}
\author{\IEEEauthorblockN{Robin E Yancey}
\IEEEauthorblockA{Department of\\Computer Science\\
University of California, Davis\\
Davis, California 95616\\
Email: reyancey@ucdavis.edu}}

\maketitle

\begin{abstract}

With technological advances leading to an increase in mechanisms for
image tampering, fraud detection methods must continue to be upgraded to
match their sophistication. One problem with current methods is that
they require prior knowledge of the method of forgery in order to
determine which features to extract from the image to localize the
region of interest. When a machine learning algorithm is used to learn
different types of tampering from a large set of various image types,
with a large enough database we can easily classify which images are
tampered. However, we still are left with the question of which features to train on, and how to localize the manipulation. In this work, deep learning for object detection is adapted to tampering detection to solve these two problems, while fusing features from multiple classic techniques for improved accuracy. A Multi-stream version of the Faster RCNN network will be employed with the second stream having an input of the element-wise sum of the ELA and BAG error maps to provide even higher accuracy than a single stream alone.

\end{abstract}

\begin{IEEEkeywords}
Faster RCNN, Deep learning, Image fraud
\end{IEEEkeywords}

\section{Introduction}  

Images are often trusted as evidence or proof in fields such as
journalism, forensic investigations, military intelligence, scientific
research and publications, crime detection and legal proceedings,
investigation of insurance claims, and medical imaging \cite{article1}.
In order to protect legal and political photos while maintaining
research integrity or reproducibility, image manipulation detection is a
highly necessary tool \cite{Cromey2010AvoidingTP}. As technology
advances, common image tampering techniques such as \textit{retouching} or \textit{resampling} which involves geometric transformations on part of the image, \textit{image splicing}, \textit{copy-move} fraud \cite{eval}, or \textit {removal}, are widely available to the public. Worse yet,
this often includes post-processing such as Gaussian smoothing, making
it even more difficult for humans to recognize the tampered regions with
the naked eye. Due to the difficulty of distinguishing fake and
authentic images, research in this field has become integral to
preventing hacking. 


However, detection of different methods such as copy-paste fraud, added WGN (White Gaussian Noise), and color enhancements, each require different filters and algorithms which must also be applied at different sized
bounding boxes depending on the size of the tampered region \cite{article}. 
These details are often not provided, making it it difficult to determine which technique to apply to which image. A method of detection that
is generalizable to various differences between images or even new types
of tampering is of great need today.

\subsection{CNN for Tampering Detection}

With the increase in image data available, and the increase in efficiency of modern GPUs to handle bigger problems, there has been interest in
the application of machine learning for image fraud detection. These
more sophisticated techniques have been able to train a model to 
estimate the probability of the images feature map (or sub-image feature
blocks) being tampered \cite{7045718}. Convolutional Neural Networks
(CNN)s, are particularly well suited for image tampering detection due to their ability to automatically learn a combination of highly detailed or pixel-level features, unable to be detected by the human eye. They have been shown capable of detecting textures, noise, and resampling much more efficiently  than classic techniques in a number of studies \cite{8014966} \cite{7113799}  \cite{7965621}  \cite{8456123}
\cite{8456113}.

With the increase in use of deep-learning for more and more
diverse image tasks, one study found that networks which are trained
in object-detection can be be adapted to manipulation detection. In this case, instead of localizing the objects in the image, the network can be used to localize
tampering artifacts by re-training on the manipulated dataset so that the network learns the manipulated features. Using a network designed for extracting regions of interest solves the problem of having to apply the specified network or filter to each sub-image box. For example, Zhou et. al. \cite{Zhou2018LearningRF} employed Faster RCNN to out-perform the speed and accuracy of image forgery detection over all previous classic methods and most CNN-based on multiple popular image datasets. A bilinear
approach was then used to simultaneously examine both the RGB image content and
noise information, providing an even \textit{higher} accuracy. This helped by combining some of the specific noise information extracted from the image with those picked up by the CNN.

\subsection{The Present Work}

However, comprehensive experimental results on multiple datasets have shown that our version of Error Level Analysis (ELA) and Block Artifact Grid (BAG) method work much better than various Noise Analysis (NA), DCT-based, or PCA-based methods in extracting tampering artifacts not brought out by the CNN alone \footnote{Test codes and example output images can be found at \cite{imagefraud}.}. This is because ELA simultaneously extracts the change in the compression local noise artifacts. Further, when the error level output map is combined with the Block Artifact Grid (BAG) method map (eg. summed pixel-wise) results are even more superior over other top classic methods.

These results have also verified that this algorithm will work on any image type since all images will have different levels of compression, regardless of whether they were originally store din JPEG. The BAG method was proven effective on detecting both copy-move and splicing forgery with varying levels of compression or quality level. This was shown by Wang et al. \cite{1038064} and Liu et. al. \cite{JPEGproof}, which tested compression rates of 5 \% to 100 \% including those with added noise.

\section{Methods \& Materials}

\begin{figure}[htbp]
\centerline{\includegraphics[scale=0.2]{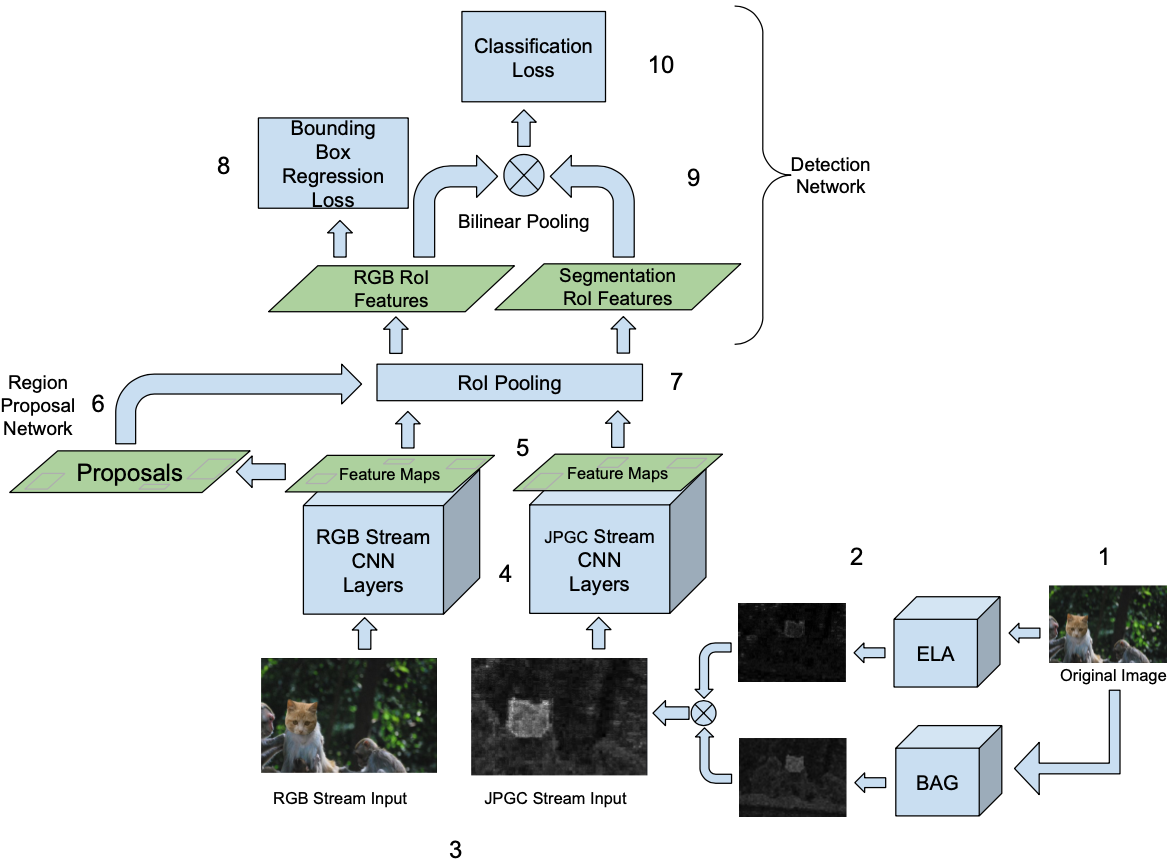}}
\caption{Multi-Stream Faster R-CNN Block diagram: 1. The original image is input into each filter 2. Both output error maps are combined 3. The original image and the output of ELA/BAG are each input into a separate stream of CNN 4. RGB stream and JPGC stream 5. The last layer of the CNN outputs anchors with multiple scales and aspect ratios which are used for the RPN to propose regions. 6. Only the last layer of the RGB stream CNN is used as input to the RPN, so both streams share these region proposals. 7. The ROI pooling layer selects spatial features from each stream and outputs a fixed length feature vector for each proposal. 8. The RoI features from the RGB stream alone are used for the final bounding box location prediction. 9. Bilinear Pooling is used to obtain spatial co-occurrence features from both streams. 10. The final predicted classes are output from the FCN and soft-max layers using sparse cross entropy loss. }
\label{diagram}
\end{figure}

The Multi-Stream Faster R-CNN framework presented here builds on the Faster RCNN network
\cite{7485869} and is a modification of the bilinear Faster RCNN \cite{Zhou2018LearningRF}. The JPEG compression
stream will have an input of the combined BAG and ELA maps of the image to provide additional features of manipulation as shown in \ref{diagram}.

\paragraph{Multi-stream Faster RCNN}

\subsection{JPEG Compression (JPGC) Stream}

Since higher error levels produce higher numbered pixels in the output maps from BAG and ELA, the element-wise sum of the two has the highest numbered (brightest white) pixels. This makes a more defined localized tampered region. The following two sections provide more detail on each individual method.

\subsubsection{Block Artifact Grid (BAG)}
The Block Artifact Grid (BAG) method uses the difference in the JPEG Quality (Q levels) found throughout image blocks to estimate the locations with high amounts of artifacts indicated by different compression rates \cite{4284574}.  The steps of the BAG method (which are similar to reverse JPEG compression) are summarized below:
\begin{itemize}
\item Divide the image into $8\times8$ blocks. Take the DCT of the blocks (using an $8\times8$ DCT matrix and matrix multiply).
\item Make a histogram of the color-quantized DCT values for each of the 64 locations of the blocks (where the number of blocks and the number of values in each histogram is equal to the number that can fit into the image).
\item Take the Fast-Fourier Transform (FFT) of the histogram of each of the 64 frequencies to get the periodicity and then power spectrum to get peaks.
\item Calculate the number of local minimums of the extrema. This is the estimated Q value.
\item Get a Q estimate for at least 32 Q values, and use it to calculate the block artifact (error in the Q value) for each image block. Output an error map of the image.
%
\end{itemize}
%

%
\subsubsection{Error Level Analysis (ELA)}

The Error Level Analysis (ELA) outputs is an image that
is created as follows:  One saves the image at a slightly lower JPEG
Q level, reads it back in, and computes the pixel-by-pixel
difference within $8 \times 8$ blocks from the original image. Since image
regions with lower Quality in the original image will degrade at a
higher rate when compressed, subtracting the decompressed image from the
original image gives the difference in Q levels in each block.
Image blocks that originally had lower Quality levels will have the
highest error and brightest color in the output.







\subsection{Bilinear Pooling}

Since the RGB stream alone has been shown to be highly accurate in
detection of manipulated regions, only this stream provides the region proposals of the RPN layer \cite{Zhou2018LearningRF}. Bilinear pooling is used to combine both streams while maintaining the spatial information. The output is $x = f_{RGB}^{T}$ $f_{JPGC}$, where $f_{RGB}$ is the RoI of the RGB stream and $f_{JPGC}$ is the RoI of the JPEG compression analysis stream. The total loss function is the sum of all of the RPN, fused classification, and regression losses, as shown in Equation \ref{eq2}:


\begin{equation}
L_{total} = \\
L_{RPN} + L_{tamper} (f_{RGB}, f_{JPGC}) + L_{bbox}(f_{RGB})  \label{eq2}
\end{equation}

where

\begin{itemize}
\item $L_{RPN}$ denotes the RPN loss 
\item $L_{tamper}$ denotes the final cross entropy classification loss (based on the output of multi-stream pooling) 
\item $L_{bbox}$ denotes the final bounding box regression loss
\item $f_{RGB}$ represents the RoI from the RGB stream
\item $f_{JPGC}$ represents the RoI from the JPEG compression stream 
\end{itemize}

\subsection{Experimental Setup}

\subsubsection{Hardware}
The model was implemented in Python with a modified version of the official Faster RCNN library \cite{renNIPS15fasterrcnn} on a Quadro 6000 cloud GPU. 

\subsubsection{Datasets}
The network was first trained and tested on a self-made spliced
image database constructed from the PASCAL VOC data \cite{pascal-voc-2012}. It was created by digitally selecting the random objects by their pixel maps provided in each dataset, pasting them into another image, and moving the new object annotation with it. Second, it was fine-tuned and tested on a few classic image manipulation datasets, which have been highly re-used in literature. These are below.

\begin{itemize}

\item CASIA 1 \& 2 (2013) \cite{6625374}: sizes $374 \times 256$, $320 \times 240$ to $800 \times 600$; Includes splicing with pre/post-processing, in TIFF/JPEG/BMP image formats.

\item CoMoFoD (2013) \cite{comofod}: sizes  $512 \times 512$; Includes copy-move, in JPEG/PNG format.

\item COVER (2016) \cite{wen2016}: sizes $400 \times 486$; Includes copy-move forged images.

\end{itemize}

All non-JPEG images were first converted to JPEG. Also, only images in the CASIA datasets with clear bounding boxes were used for testing so that it would be easy to make a fair judgement of the overall prediction accuracy on this dataset. Table \ref{tab1} provides the Test/Train distributions and number of Training steps used respectively, for each dataset.

\begin{table}[!t]
\renewcommand{\arraystretch}{1.3}
\caption{Train/Test Data Setups}
\label{tab1}
\centering
\begin{tabular}{|c|c|c|c|}
\hline
Dataset & Test/Train & Train Steps \\ \hline 
Synthetic & 5k/5k & 45k \\ 
COVER  & 10/100 & 15k  \\ 
CASIA & 50/2886 & 25k \\ 
CoMoFoD & 15/143 & 20k \\ 
\hline
\end{tabular}
\end{table}

\subsubsection{Accuracy Calculation}

The results are based on the F-score as in \ref{1}, which is a harmonic mean of precision and recall (sensitivity):

\begin{equation}  F-score = \frac{2 \cdot (precision \cdot sensitivity) }{(precision + sensitivity)} \end{equation} \label{1}

The precision measures how accurate the predictions are using the percentage of the correct predictions out of the total. 
It is calculated using the $FP$ which represents the number of false positives, and $TP$ which is the number of true positives, as in Equation \ref{eq-p}:
\begin{equation} Precision = \frac{TP}{FP+TP} \label{eq-p} \end{equation} The recall measures how well positives are found, where $FN$ is the number of false negatives (those ground truths which were not detected), as in Equation \ref{eq-r} \begin{equation} Recall = \frac{TP}{FN+TP} \label{eq-r} \end{equation}

\section{Results}

\subsection{Synthetic Dataset Tests}

The accuracy was over 90\% on the synthetic dataset. Top methods such as, Yan et. al. \cite{pvtest} also created a synthetic dataset using PASCAL VOC 2012 and obtained 87\%. They used a CNN-based deep architecture, which consists of three feature extraction blocks and a feature fusion module.

\subsection{Official Image Manipulation Dataset Tests}

\begin{figure}[htbp]
\centerline{\includegraphics[scale=0.17]{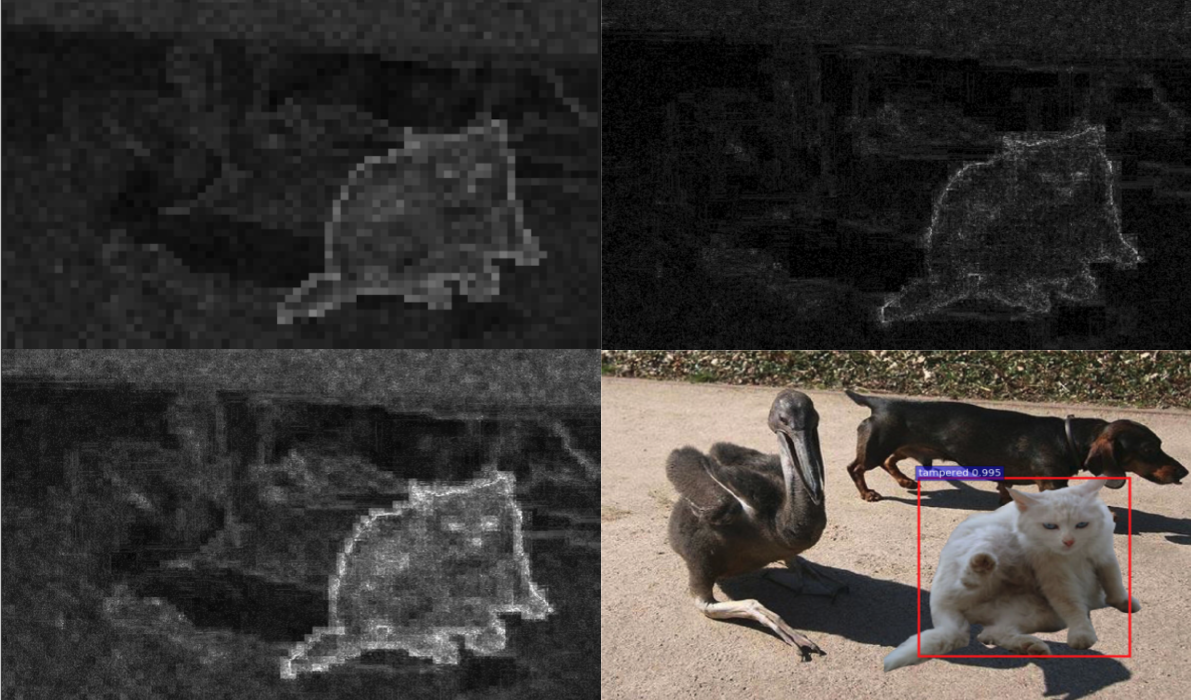}}
 \caption{Top Left: Output of BAG, Top Right: Output of ELA, Bottom Left: Sum of the Outputs, Bottom Right: Output of the MS-Faster RCNN Model}
 \label{compare}
 \end{figure}

Figure \ref{compare} shows the output of BAG and ELA in the top left and right, respectively. The bottom row shows the sum of the outputs and the output of the MS-Faster RCNN, respectively. The predicted bounding box (in red) is around the tampered region (the cat).

Table \ref{tab2} shows the performance comparison to the single stream version (RGB Net) and the bilinear version with the second stream as the noise features (RGB-N) \cite{Zhou2018LearningRF}, as well as the BusterNet \cite{Wu2018BusterNetDC}, which is th past top performing model on these datasets. Kumar and Bhavasar obtained slightly higher scores on CASIA and CoMoFoD dataset to 80.3 \% and 78.8\%, but only tested the copy-paste forged images.

\begin{table}[!t]
\renewcommand{\arraystretch}{1.3}
\caption{Performance Comparison to Other Models}
\label{tab2}
\begin{tabular}{|c|c|c|c|c|}
\hline
Dataset & MS-FRCNN & RGB-N & RGB Net & BusterNet \\ \hline 
COVER  & 0.82 & 0.437 & 0.39 & -  \\ 
CASIA & 0.69 & 0.408 & 0.39 & 0.67  \\ 
CoMoFoD & 0.59 & - & - & 0.49  \\ 
\hline
\end{tabular}
\end{table}

\section{Conclusion}

The multi-stream version localized tampered regions better than other methods. Fusing more classic features including the JPEG or compression artifacts helps extract more tampering types than other features such as noise. Older methods normally use a sliding window of feature maps to test for manipulated regions, while a deep network uses bounding box regression on various anchor box sizes to estimate the probability of a region being tampered allowing us to automatically capture more information. As future work, fusing more classically used features, obtained from DCT or PCA-based methods may help to even further increase the score.
%
%
%

\newpage
\bibliography{MLSP} {}
\bibliographystyle{IEEEtran}

\end{document}